\documentclass[10pt]{article}

\usepackage{natbib} 
\usepackage[final]{tackling_climate_workshop_style}
\usepackage{graphicx}
\usepackage{subcaption}
\usepackage{float}
\usepackage{booktabs}
\usepackage{hyperref}
\usepackage{url}
\usepackage{multirow}
\usepackage{amsmath}

\title{Spatial Uncertainty Quantification in Wildfire Forecasting for Climate-Resilient Emergency Planning}

\author{Aditya Chakravarty \\
Independent Research \\
San Francisco, CA \\
\texttt{chakravarty.aditya28@gmail.com}}

\begin{document}

\maketitle

\begin{abstract}
Climate change is intensifying wildfire risks globally, making reliable forecasting critical for adaptation strategies. While machine learning shows promise for wildfire prediction from Earth observation data, current approaches lack uncertainty quantification essential for risk-aware decision making. We present the first systematic analysis of spatial uncertainty in wildfire spread forecasting using multimodal Earth observation inputs. We demonstrate that predictive uncertainty exhibits coherent spatial structure concentrated near fire perimeters. Our novel distance metric reveals high-uncertainty regions form consistent 20–60 meter buffer zones around predicted firelines—directly applicable for emergency planning. Feature attribution identifies vegetation health and fire activity as primary uncertainty drivers. This work enables more robust wildfire management systems supporting communities adapting to increasing fire risk under climate change.

\end{abstract}

\section{Introduction}

Wildfires have become an escalating global crisis, intensified by climate change, prolonged droughts, and expanding human development. Most recently in January 2025, Southern California experienced wildfire events that have been among the costliest natural disasters in U.S. history. Wildfires in the European Union have become increasingly frequent and severe, with over 166,000 hectares burned by May 2025, nearly three times the long-term average, driven by climate change and affecting regions beyond the traditional Mediterranean hotspots~\citep{effis2025}. Globally, regions such as the Amazon, North America, Australia, and parts of Africa have witnessed unprecedented wildfire activity, leading to significant ecological damage, loss of biodiversity, and adverse health effects due to smoke exposure \citep{copernicus2024}.

As climate change continues to exacerbate wildfire risks, there is an urgent need for accurate, high-resolution wildfire forecasting to aid in early response, resource allocation, and risk mitigation. While remote sensing products like VIIRS \citep{schroeder2014viirs} and MODIS provide near real-time fire detections, they do not forecast how wildfires will evolve in the days to come.

Machine learning has emerged as a scalable alternative for fire forecasting \citep{radke2019firecast, bolt2022spatiotemporal}, learning directly from remote sensing and historical fire data. Recent work has demonstrated strong predictive performance using multimodal remote sensing inputs, but critically lacks uncertainty quantification—leaving users without insight into where or why models might fail.

Despite the operational risks involved, no prior work has investigated uncertainty quantification in high-resolution wildfire forecasting. Most existing approaches are fully deterministic, producing binary or probabilistic predictions without expressing model confidence. This omission is especially concerning for wildfire response, where uncertainty-aware decision-making is essential for frontline planning, containment strategies, and risk assessment.

\section{Climate Impact and Pathway}

\textbf{Direct Climate Relevance:} This work directly addresses climate change impacts through improved wildfire forecasting capabilities. Wildfires are a key climate-related hazard that will increase in frequency and severity under warming scenarios. Better prediction and uncertainty quantification enables more effective climate adaptation strategies.

\textbf{Pathway to Impact:} Our uncertainty-aware wildfire forecasting system provides actionable insights for multiple stakeholders: (1) Emergency managers can use spatial uncertainty maps to optimize resource allocation and evacuation planning; (2) Fire suppression teams benefit from buffer zone estimates that indicate where model predictions are most reliable; (3) Climate adaptation planners can incorporate uncertainty estimates into long-term risk assessments and land-use planning decisions.

\textbf{Adaptation Focus:} This work primarily supports climate adaptation by improving society's ability to respond to increased wildfire risk. The spatial uncertainty framework helps communities and agencies better prepare for and respond to fire events in a changing climate.

\textbf{Operational Integration:} The 20–60 meter uncertainty buffer zones we identify correspond to tactically relevant scales for incident management teams. These estimates can be directly integrated into existing fire management workflows and decision support systems.

\section{Methods}

\subsection{Dataset}
We conduct all experiments on the publicly available WildfireSpreadTS dataset\footnote{\url{https://github.com/SebastianGer/WildfireSpreadTS}}~\citep{gerard2023wildfirespreadts}, which provides spatial-temporal cubes of 64×64 patches centered on active wildfire regions. Each sample consists of 5 days of multimodal input features (Sentinel-2 reflectance bands, meteorological variables, NDVI, slope, and other static features), and a binary burn mask for a future day as target. The dataset includes 607 wildfire events across the western United States from January 2018 to October 2021, with a total of 13,607 daily images spanning diverse ecosystems and terrain.

\subsection{Model Architecture}
We use the UTAE model \citep{garnot2020satellite}, a transformer-based spatiotemporal encoder-decoder architecture designed for multitemporal satellite image time series. UTAE has shown strong performance on change detection and land cover segmentation tasks using Sentinel-2 data, and is well-suited for wildfire spread forecasting where temporal patterns are key. The model has approximately 1M parameters, making it computationally efficient compared to larger transformer architectures while maintaining strong performance.

\subsection{Uncertainty Quantification}
We evaluate three uncertainty quantification approaches for pixel-wise wildfire spread prediction:

\textbf{Monte Carlo (MC) Dropout}: Dropout layers remain active at test time, and 20 stochastic forward passes are performed. The per-pixel mean and variance of predicted probabilities quantify epistemic uncertainty.

\textbf{Deep Ensembles}: Multiple independent UTAE models are trained with distinct random seeds, each employing MC Dropout with 20 stochastic forward passes at inference. Predictions are aggregated to quantify uncertainty from weight initialization, training stochasticity, and dropout randomness.

\textbf{Bayesian Neural Networks (BNN)}: We treat network weights as random variables with learned probability distributions using variational approximation (Bayes-by-Backprop). Multiple stochastic samples of weight distributions are drawn at inference to obtain mean wildfire burn probability and associated epistemic uncertainty maps. To assess probabilistic quality, we compute Expected Calibration Error (ECE), Brier Score, and Negative Log-Likelihood (NLL).

\section{Results}

\subsection{Feature Importance and Model Performance}
Through systematic ablation studies, we find that vegetation-based features (VIIRS bands M11, I2, I1, NDVI, and EVI2) combined with active fire indicators achieve the highest performance (AP: 0.378 ± 0.083), outperforming weather, topography, and land cover feature groups. This vegetation-focused model forms the basis for all uncertainty analyses. Figure~\ref{fig:input_modalities_example} shows the key input modalities used by the model.

\begin{table}[htbp]
\centering
\caption{Mean Average Precision (AP) across 12 folds for different feature groups using UTAE.}
\label{tab:utae_baseline_runs}
\small
\setlength{\tabcolsep}{4pt}
\begin{tabular}{l@{\hskip 6pt}c}
\toprule
\textbf{Feature Group} & \textbf{Mean AP} \\
\midrule
Persistence baseline     & $0.191 \pm 0.063$ \\
Vegetation + active fire     & $0.378 \pm 0.083$ \\
Weather + active fire        & $0.323 \pm 0.078$ \\
Land cover + active fire     & $0.319 \pm 0.092$ \\
Topography + active fire     & $0.317 \pm 0.082$ \\
All Features + active fire    & $0.319 \pm 0.077$ \\
\bottomrule
\end{tabular}
\end{table}

\subsection{Uncertainty Calibration}

\begin{table}[htbp]
\centering
\caption{Calibration metrics (12-fold averages) for the three UQ approaches. Lower values indicate better calibration. Deep Ensembles achieve the strongest calibration across all metrics.}
\label{tab:uq_comparison_revised}
\small
\setlength{\tabcolsep}{6pt}
\begin{tabular}{lccc}
\toprule
\textbf{Metric} & \textbf{MC Dropout} & \textbf{BNN} & \textbf{Deep Ensemble} \\
\midrule
ECE         & $0.536 \pm 0.015$ & $0.525 \pm 0.014$ & $\mathbf{0.512 \pm 0.018}$ \\
Brier Score & $0.294 \pm 0.012$ & $0.283 \pm 0.019$ & $\mathbf{0.265 \pm 0.009}$ \\
NLL         & $0.805 \pm 0.020$ & $0.794 \pm 0.054$ & $\mathbf{0.731 \pm 0.023}$ \\
\bottomrule
\end{tabular}
\end{table}

\subsection{Spatial Uncertainty Structure}
\begin{figure*}[!ht]
    \centering
    \includegraphics[width=0.6\linewidth]{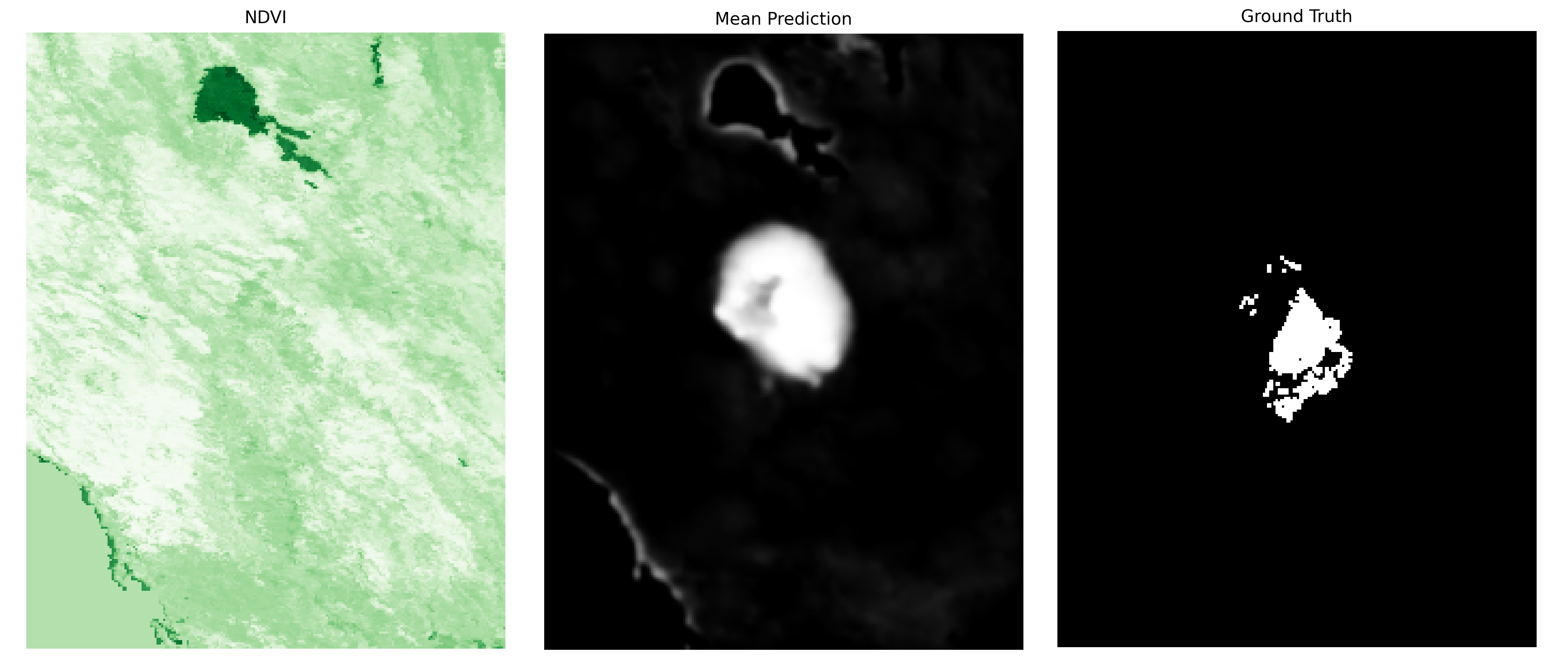}
    \vspace{0.3em}
    \includegraphics[width=0.6\linewidth]{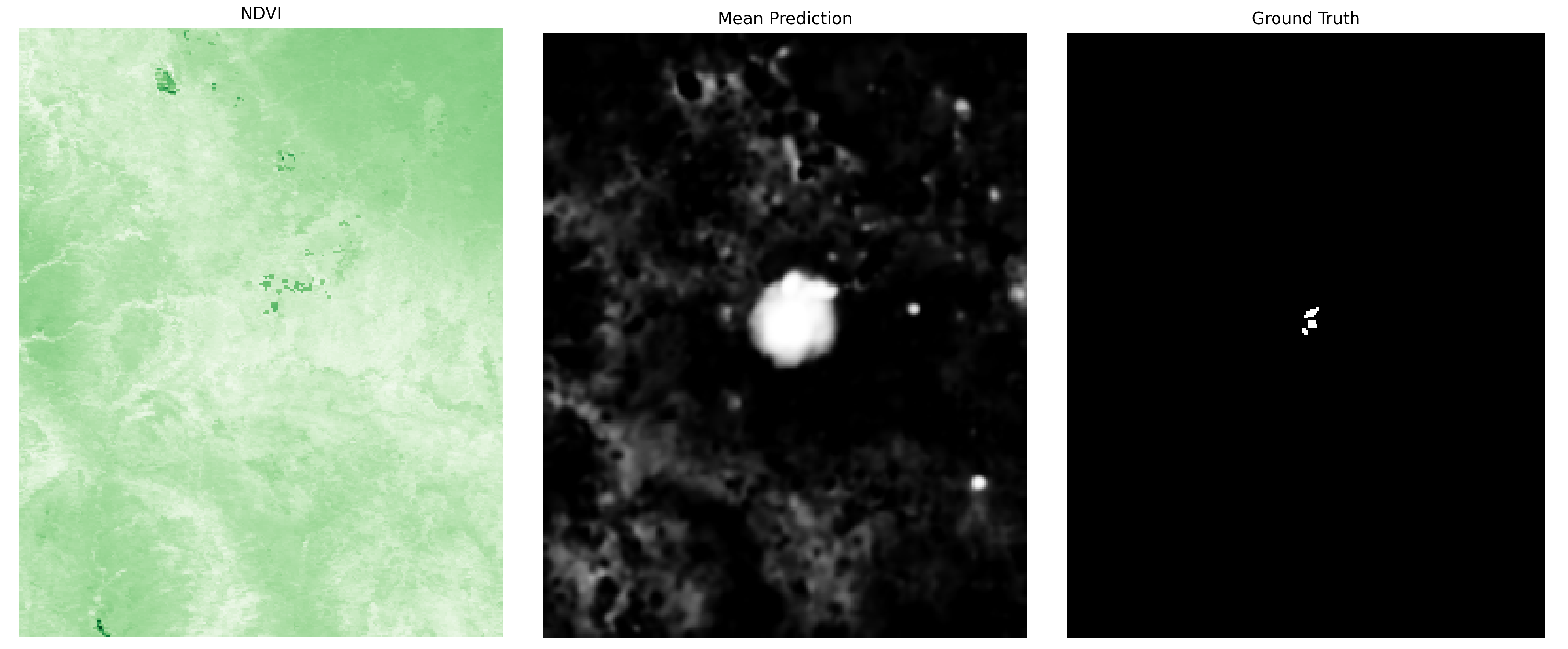}
    \caption{
        Qualitative comparison of model predictions for three fire events of varying size: large (top), medium (middle), and small (bottom).
        Each row shows the NDVI input, mean prediction from a Deep Ensemble, 
        and the ground-truth burn mask. The influence of vegetation features on the model's mean predictions is clearly evident. The events span approximately 125.6 acres (large), and 5.2 acres (small).
    }
    \label{fig:uq-gt-ndvi-overlay}
\end{figure*}

Uncertainty estimates form coherent spatial patterns concentrated near predicted fire perimeters rather than scattered noise. Figure~\ref{fig:uq-gt-ndvi-overlay} shows qualitative examples for three fire events of varying sizes, demonstrating that uncertainty is sharply localized near fire perimeters in larger fires, while appearing more diffuse in smaller or fragmented fires.

\subsection{Quantitative Buffer Zone Analysis}
We introduce a centroid-oriented boundary distance metric to quantify spatial prediction errors, illustrated in Figure~\ref{fig:centroid_schematic}. Analysis reveals consistent uncertainty buffer zones of 28–35 meters using our centroid metric, 47–64 meters using Average Surface Distance, and 148–166 meters using Hausdorff Distance (more details outlined in the appendix).

\begin{table}[h]
\centering
\caption{Most likely spatial offsets (peak distances) from KDE histograms for each distance metric, providing empirical estimates of operational buffer zones.}
\label{tab:kde_peaks_revised}
\small
\setlength{\tabcolsep}{8pt}
\begin{tabular}{lll}
\toprule
\textbf{Distance Metric} & \textbf{Feature Set} & \textbf{Peak Distance (m)} \\
\midrule
Centroid Boundary Distance & Vegetation    & 32.19 \\
                           & All Features  & 33.48 \\
\midrule
Average Surface Distance   & Vegetation    & 64.15 \\
                          & All Features  & 55.86 \\
\midrule
Hausdorff Distance        & Vegetation    & 165.78 \\
                         & All Features  & 155.67 \\
\bottomrule
\end{tabular}
\end{table}

\section{Discussion and Conclusion}

We present the first systematic analysis of spatial uncertainty in high-resolution, EO-based wildfire forecasting. Using multimodal satellite inputs and uncertainty quantification (UQ) methods, we find that epistemic uncertainty forms coherent 20--60\,m buffer zones aligned with fire perimeters and vegetation gradients. These interpretable patterns offer a practical proxy for operational planning in climate adaptation and disaster response.

Our centroid-oriented boundary distance metric captures such zones efficiently, though it may be less reliable with disconnected fire fronts or misaligned centroid axes. Feature attribution shows vegetation health and recent fire activity as the main drivers of predictive confidence.

Limitations include focus on epistemic (not aleatoric) uncertainty, U.S.-only coverage, and restricted input features; adding weather or topography reduced performance, likely due to temporal and spatial resolution mismatches.

Overall, spatial uncertainty is not noise but a signal: integrating structured UQ maps into workflows can support risk-aware decision-making, resource allocation, and safer wildfire management under climate change.


\bibliography{main}

\begin{thebibliography}{10}
\providecommand{\natexlab}[1]{#1}
\providecommand{\url}[1]{\texttt{#1}}
\expandafter\ifx\csname urlstyle\endcsname\relax
  \providecommand{\doi}[1]{doi: #1}\else
  \providecommand{\doi}{doi: \begingroup \urlstyle{rm}\Url}\fi

\bibitem[Bolt et~al.(2022)]{bolt2022spatiotemporal}
Andrew Bolt et~al.
\newblock A spatio-temporal neural network forecasting approach for emulation of firefront models.
\newblock In \emph{Signal Processing: Algorithms, Architectures, Arrangements, and Applications (SPA)}, 2022.

\bibitem[Cunningham et~al.(2024)Cunningham, Jones, and Patel]{copernicus2024}
Alex Cunningham, Sarah Jones, and Ravi Patel.
\newblock State of wildfires 2023–2024.
\newblock \emph{Earth System Science Data}, 16:\penalty0 3601--3620, 2024.

\bibitem[{EFFIS}(2025)]{effis2025}
{EFFIS}.
\newblock Wildfire season 2025: Early trends and eu response.
\newblock \url{https://effis.jrc.ec.europa.eu/}, 2025.
\newblock Accessed: 2025-05-29.

\bibitem[Garnot \& Landrieu(2021)Garnot and Landrieu]{garnot2020satellite}
Vivien Sainte~Fare Garnot and Loic Landrieu.
\newblock Panoptic segmentation of satellite image time series with convolutional temporal attention networks.
\newblock In \emph{Proceedings of the IEEE/CVF International Conference on Computer Vision (ICCV)}, 2021.

\bibitem[Gerard et~al.(2023)Gerard, Zhao, and Sullivan]{gerard2023wildfirespreadts}
Sebastian Gerard, Yu~Zhao, and Josephine Sullivan.
\newblock Wildfirespread{TS}: A dataset of multi-modal time series for wildfire spread prediction.
\newblock In \emph{Thirty-seventh Conference on Neural Information Processing Systems Datasets and Benchmarks Track}, 2023.
\newblock URL \url{https://openreview.net/forum?id=RgdGkPRQ03}.

\bibitem[Lahrichi et~al.(2025)Lahrichi, Johnson, and Malof]{lahrichi2025predicting}
Saad Lahrichi, Jesse Johnson, and Jordan Malof.
\newblock Predicting next-day wildfire spread with time series and attention.
\newblock \emph{arXiv preprint arXiv:2502.12003}, 2025.

\bibitem[Liu et~al.(2021)Liu, Lin, Cao, Hu, Wei, Zhang, Lin, and Guo]{liu2021swintransformerhierarchicalvision}
Ze~Liu, Yutong Lin, Yue Cao, Han Hu, Yixuan Wei, Zheng Zhang, Stephen Lin, and Baining Guo.
\newblock Swin transformer: Hierarchical vision transformer using shifted windows.
\newblock 2021.

\bibitem[Loshchilov \& Hutter(2019)Loshchilov and Hutter]{loshchilov2019decoupled}
Ilya Loshchilov and Frank Hutter.
\newblock Decoupled weight decay regularization.
\newblock In \emph{International Conference on Learning Representations (ICLR)}, 2019.
\newblock URL \url{https://openreview.net/forum?id=Bkg6RiCqY7}.

\bibitem[Radke et~al.(2019)Radke, Hessler, and Ellsworth]{radke2019firecast}
David Radke, Anna Hessler, and Dan Ellsworth.
\newblock Firecast: Leveraging deep learning to predict wildfire spread.
\newblock In \emph{IJCAI}, pp.\  4575--4581, 2019.

\bibitem[Schroeder et~al.(2014)Schroeder, Oliva, Giglio, and Csiszar]{schroeder2014viirs}
Wilfrid Schroeder, Patricia Oliva, Louis Giglio, and Ivan~A Csiszar.
\newblock The new viirs 375 m active fire detection data product: Algorithm description and initial assessment.
\newblock \emph{Remote Sensing of Environment}, 143:\penalty0 85--96, 2014.

\end{thebibliography}
\bibliographystyle{tmlr}

\newpage
\appendix
\section{Appendix}

\begin{figure*}[htbp]
    \centering
    \includegraphics[width=0.8\linewidth]{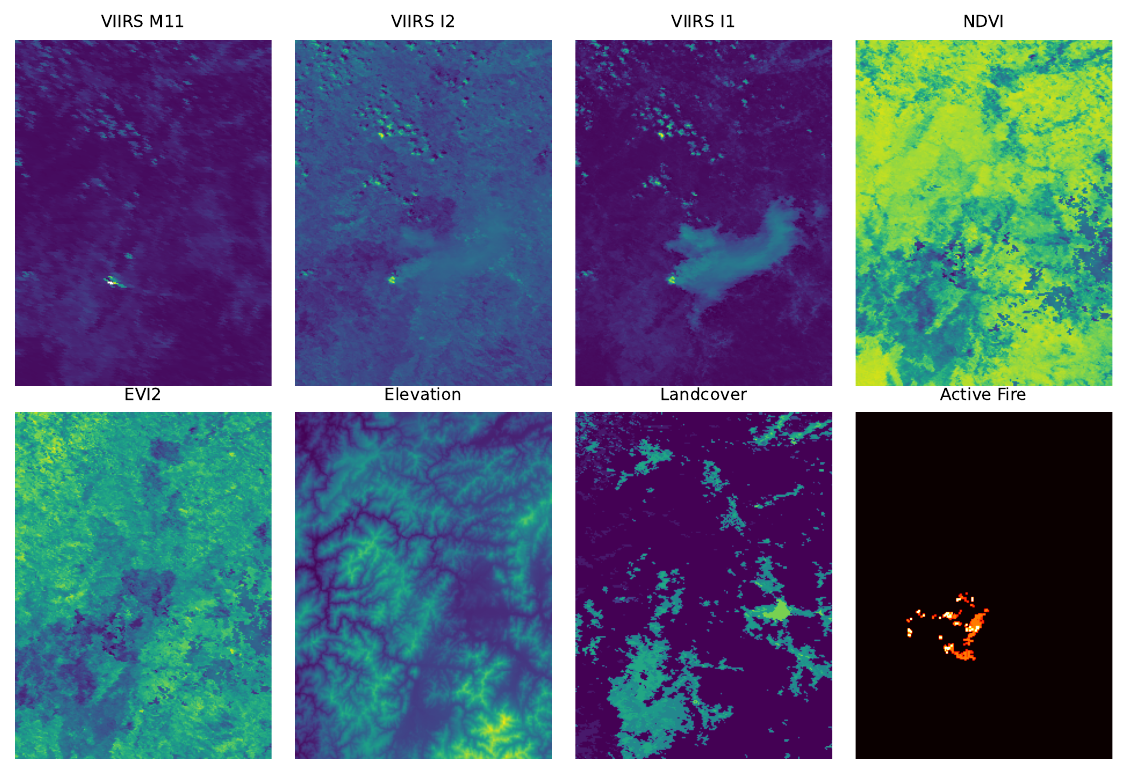}
    \caption{
        Example input channels from a single sample at prediction time, including Sentinel-2 bands, NDVI, EVI2, and active fire features. These inputs are provided as a 5-day sequence to the model.
    }
    \label{fig:input_modalities_example}
\end{figure*}

\subsection{Feature Attribution}
\begin{figure}[htbp]
    \centering
    \includegraphics[width=0.6\linewidth]{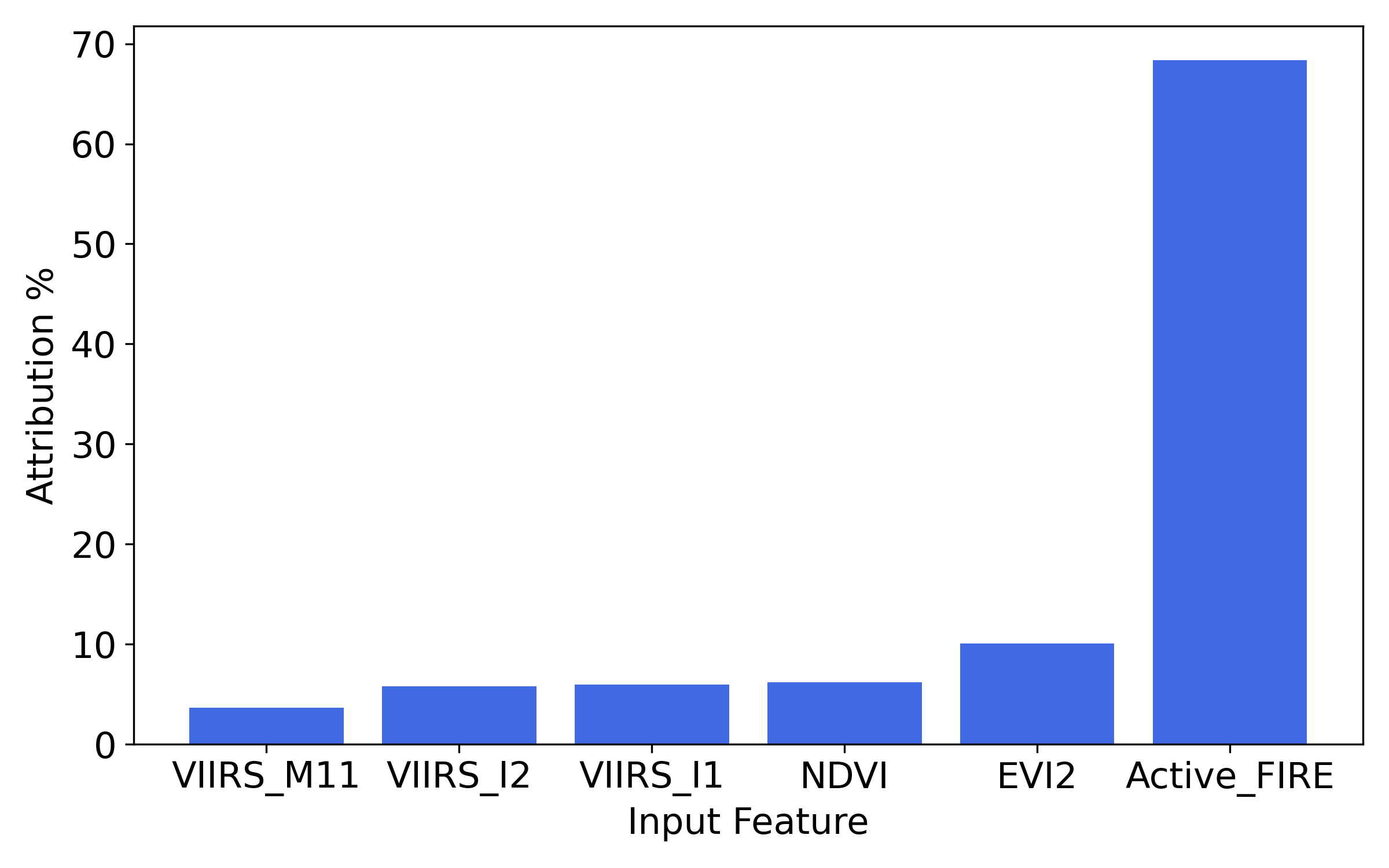}
    \caption{
        Feature importance scores computed using Integrated Gradients. Active fire presence dominates attribution, followed by vegetation indices (NDVI, EVI2). Thermal bands are less influential.
    }
    \label{fig:gradient-feature-attribute}
\end{figure}

Using Integrated Gradients on a CNN surrogate model (R² = 0.81 fidelity), we find that recent fire activity and vegetation indices (NDVI, EVI2) dominate predictive influence, with lower contributions from thermal reflectance bands (Figure~\ref{fig:gradient-feature-attribute}).

\subsection{UTAE Model Selection Rationale and Procedural Details}

We provide detailed implementation information for the UTAE baseline, including architectural, optimization, and data processing details. We also motivate our choice of UTAE over alternative temporal architectures.

\paragraph{Model Architecture.}
We adopt the UTAE architecture\citep{garnot2020satellite}, a U-Net variant with a Temporal Attention Encoder that applies simplified multi-head self-attention across the temporal dimension at the bottleneck. These temporal attention weights are up sampled and applied to the skip connections, enabling dynamic selection of temporally relevant features. The model has approximately 1.1M parameters and includes a dropout rate of 0.1 after each attention block to prevent overfitting.

\paragraph{Motivation for UTAE.}
We chose UTAE for its lightweight parameter count, proven effectiveness on spatiotemporal satellite time series segmentation, and its compatibility with variable-length sequences. Although newer transformer-based architectures, such as Swin Transformers \citep{liu2021swintransformerhierarchicalvision}, have been explored for spatiotemporal wildfire modeling \citep{lahrichi2025predicting}, empirical evidence suggests that they do not outperform UTAE in next-day wildfire spread prediction. On the downside, these larger architectures contain roughly 27M parameters compared to UTAE's lightweight 1M parameters, requiring significantly more computational resources for both pretraining and fine-tuning. They are also prone to overfitting and typically demand much larger datasets to generalize effectively—an unrealistic requirement given the size and variability of current wildfire datasets. For these reasons, and given UTAE’s proven reliability and favorable trade-off between performance, computational efficiency, and robustness, we adopt UTAE as our primary model. Our empirical findings also show that UTAE outperforms ConvLSTM and standard U-Net by a margin of up to 3.9 AP points on this dataset.

\paragraph{Input Configuration.}
Each input sequence consists of 5 days of observations, each containing:
\begin{itemize}
    \item \textbf{Vegetation}: VIIRS reflectance bands (I1, I2, M11), NDVI, EVI2
    \item \textbf{Weather and Forecasts}: Precipitation, temperature (min, max), wind (speed, direction), specific humidity, PDSI, ERC, GFS forecasts of the same
    \item \textbf{Topography}: Slope, aspect, elevation
    \item \textbf{Land cover}: One-hot encoded MODIS IGBP class
    \item \textbf{Fire masks}: Timestamped detection map and binary mask
    \item \textbf{Day-of-year}: Integer mapped to temporal embedding
\end{itemize}

\paragraph{Preprocessing}
All input features are resampled to a spatial resolution of 375\,m and a temporal resolution of 24 hours. Numerical features are standardized to zero mean and unit variance, excluding angular features (sine-transformed) and categorical/binary maps. Missing values are replaced with zero. We apply the following augmentations:
\begin{itemize}
    \item Random crop to $128 \times 128$ pixels, with oversampling based on fire presence
    \item Horizontal and vertical flips, $90^\circ$ rotations
    \item Angle-aware adjustment for wind direction and aspect post-rotation
\end{itemize}
At test time, we apply center cropping to a size divisible by 32 to meet U-Net alignment constraints.

\paragraph{Optimization}
We use the AdamW optimizer~\cite{loshchilov2019decoupled} with parameters $\beta_1 = 0.9$, $\beta_2 = 0.999$, learning rate $= 0.01$, and weight decay $\lambda = 0.01$. Dropout (0.1) is applied at each temporal attention block. We use a weighted binary cross-entropy loss for loss function. The model is trained for 10,000 steps. Batch size is 32 during training and 1 during testing.

\subsection{Centroid-Aligned Boundary Distance as a Proxy for Fireline Uncertainty.}
\begin{figure}[htbp]
    \centering
    \includegraphics[width=0.7\linewidth]{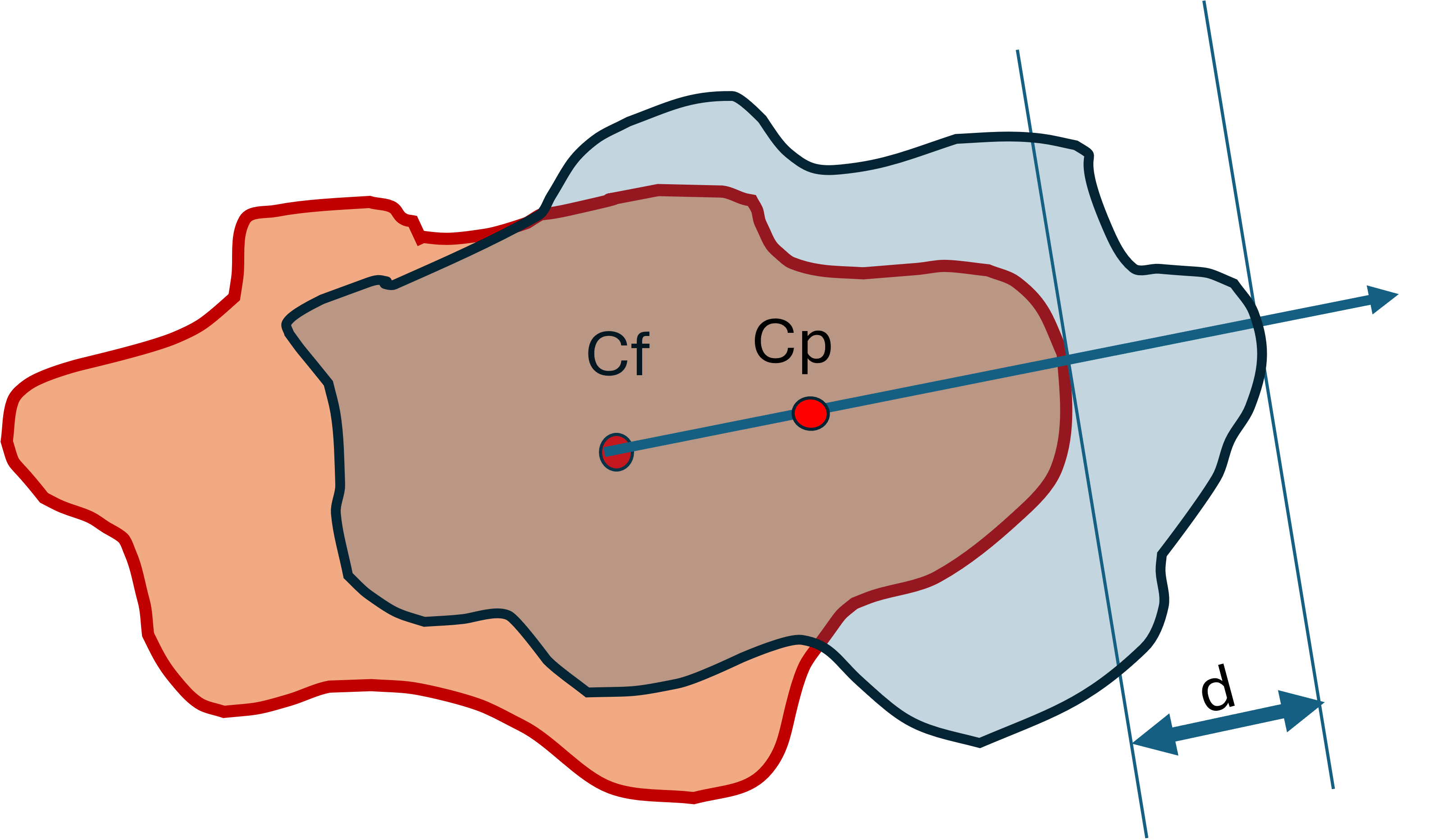}
    \caption{
        Schematic of boundary distance computation between predicted and ground truth fire masks.
    }
    \label{fig:centroid_schematic}
\end{figure}

To better understand the spatial structure of false positives in fireline predictions, we introduce a \textit{centroid-aligned boundary distance} metric. For each test instance, we compute the centroid of the ground truth burn mask ($C_f$) and the centroid of the predicted fireline ($C_p$), using the mean prediction from deep ensembled outputs thresholded at 0.95. We then trace a straight line between $C_f$ and $C_p$, and identify the nearest points along this axis where the predicted and ground truth firelines terminate. The distance between these two edge points serves as a localized estimate of spatial prediction error. We formally define the distance metric: 

Let $M_{\text{gt}}$ and $M_{\text{pred}}$ be the binary masks for the ground truth and predicted fire regions, respectively. Let $C_{f} = (x_{f}, y_{f})$ and $C_{p} = (x_{p}, y_{p})$ denote the centroids of the ground truth region and the false positive region defined as
\[
M_{\text{fp}} = M_{\text{pred}} \land \lnot M_{\text{gt}}.
\]
We define the centroid-to-centroid axis as the discrete line segment connecting $C_{f}$ and $C_{p}$, denoted by $\mathcal{L}(C_{f}, C_{p})$.
Let $\partial M_{\text{gt}}$ and $\partial M_{\text{fp}}$ denote the boundary pixels of the ground truth and false positive regions, respectively. These are computed as:
\[
\partial M = \text{dilate}(M) \land \lnot M.
\]
We identify the first boundary pixel $p_{\text{gt}} \in \partial M_{\text{gt}}$ along $\mathcal{L}(C_{f}, C_{p})$ starting from $C_{f}$, and the first pixel $p_{\text{fp}} \in \partial M_{\text{fp}}$ from the opposite direction. The centroid-oriented boundary distance $d$ is defined as:
\[
d = \| p_{\text{gt}} - p_{\text{fp}} \|_2 \cdot s,
\]
where $s$ is the pixel resolution.   Next, we compare this metric with two established distance measures: Average Surface Distance (ASD) and Hausdorff Distance (HD).

\textbf{Average Surface Distance (ASD):}
Let $D_{\text{gt}}$ and $D_{\text{fp}}$ be the distance transform maps of the complement regions $\lnot M_{\text{gt}}$ and $\lnot M_{\text{fp}}$, respectively. For each boundary pixel $p \in \partial M_{\text{gt}}$, we compute its distance to the nearest boundary pixel in $\partial M_{\text{fp}}$ using $D_{\text{fp}}[p]$. Similarly, for each boundary pixel $q \in \partial M_{\text{fp}}$, we compute its distance to the nearest boundary pixel in $\partial M_{\text{gt}}$ using $D_{\text{gt}}[q]$. The average surface distance $d_{\text{ASD}}$ is defined as:
\[
d_{\text{ASD}} = \frac{1}{2}\left(\frac{1}{|\partial M_{\text{gt}}|}\sum_{p \in \partial M_{\text{gt}}} D_{\text{fp}}[p] + \frac{1}{|\partial M_{\text{fp}}|}\sum_{q \in \partial M_{\text{fp}}} D_{\text{gt}}[q]\right) \cdot s.
\]

\textbf{Hausdorff Distance:}
Let $P_{\text{gt}} = \{p : p \in \partial M_{\text{gt}}\}$ and $P_{\text{fp}} = \{q : q \in \partial M_{\text{fp}}\}$ be the sets of boundary pixel coordinates. The directed Hausdorff distances are defined as:
\[
h(P_{\text{gt}}, P_{\text{fp}}) = \max_{p \in P_{\text{gt}}} \min_{q \in P_{\text{fp}}} \|p - q\|_2
\]
\[
h(P_{\text{fp}}, P_{\text{gt}}) = \max_{q \in P_{\text{fp}}} \min_{p \in P_{\text{gt}}} \|q - p\|_2
\]
The Hausdorff distance $d_{\text{HD}}$ is defined as:
\[
d_{\text{HD}} = \max\{h(P_{\text{gt}}, P_{\text{fp}}), h(P_{\text{fp}}, P_{\text{gt}})\} \cdot s.
\]
\paragraph{Comparative Analysis of Distance Metrics}

The three metrics offer complementary perspectives on spatial prediction errors (Table~\ref{tab:kde_peaks_revised}). The centroid-oriented boundary distance provides a directionally-informed, single-value summary that captures primary spatial offset efficiently but is limited to one spatial dimension and may miss perpendicular boundary irregularities. The Average Surface Distance (ASD) offers a more comprehensive, symmetric assessment by averaging all boundary distances, providing statistically robust summaries less sensitive to outliers, though it can mask significant local deviations by averaging them with smaller errors. The Hausdorff Distance captures worst-case spatial errors through maximum boundary separation, providing upper bound guarantees valuable for critical applications, but is highly sensitive to noise and isolated mispredictions that may not represent systematic bias. Computationally, the centroid method requires minimal resources, ASD involves distance transforms across boundary pixels, while Hausdorff requires more expensive pairwise calculations. For operational fire management, these three metrics collectively enable rapid spatial offset assessment, balanced characterization of boundary discrepancies, and identification of critical failure cases.

\end{document}